\documentclass[runningheads]{llncs}

\usepackage[dvipsnames, table]{xcolor}

\definecolor{lgreen}{RGB}{236, 255, 201}
\definecolor{nvgreen}{RGB}{118, 185, 0}
\definecolor{cvprblue}{rgb}{0.21,0.49,0.74}
\definecolor{lgray}{RGB}{245,245,245}
\usepackage{amsmath}
\usepackage{pifont}
\usepackage{float}
\usepackage{arydshln}
\usepackage{algorithm}
\usepackage{algorithmicx}
\usepackage{algpseudocode}
\usepackage{subcaption}
\algrenewcommand\alglinenumber[1]{\tiny #1:}
\DeclareMathOperator*{\argmax}{arg\,max}

\usepackage{eccv}

\usepackage{eccvabbrv}

\usepackage{graphicx}
\usepackage{booktabs}
\newcommand{\Hquad}{\hspace{0.5em}} 
\newcommand\method{MDP}
\usepackage[accsupp]{axessibility}  

\usepackage[pagebackref,breaklinks,colorlinks,citecolor=eccvblue]{hyperref}

\usepackage{orcidlink}

\begin{document}

\title{Multi-Dimensional Pruning: Joint Channel, Layer and Block Pruning with Latency Constraint}

\titlerunning{Multi-Dimensional Pruning}

\author{
Xinglong Sun\inst{1,2}\thanks{Performed during an internship at NVIDIA} \and
Barath Lakshmanan\inst{1} \and
Maying Shen\inst{1} \and
Shiyi Lan\inst{1} \and
Jingde Chen\inst{1} \and
Jose Alvarez\inst{1} 
}

\authorrunning{X.Sun et al.}

\institute{NVIDIA\and
Stanford University\\
\email{\{xinglongs, blakshmanan, mshen, shiyil, joshchen, josea\}@nvidia.com}\\
}

\maketitle

\begin{figure}[h!]
\begin{subfigure}[b]{.5\textwidth}
\begin{center}
   \includegraphics[width=.9\linewidth]{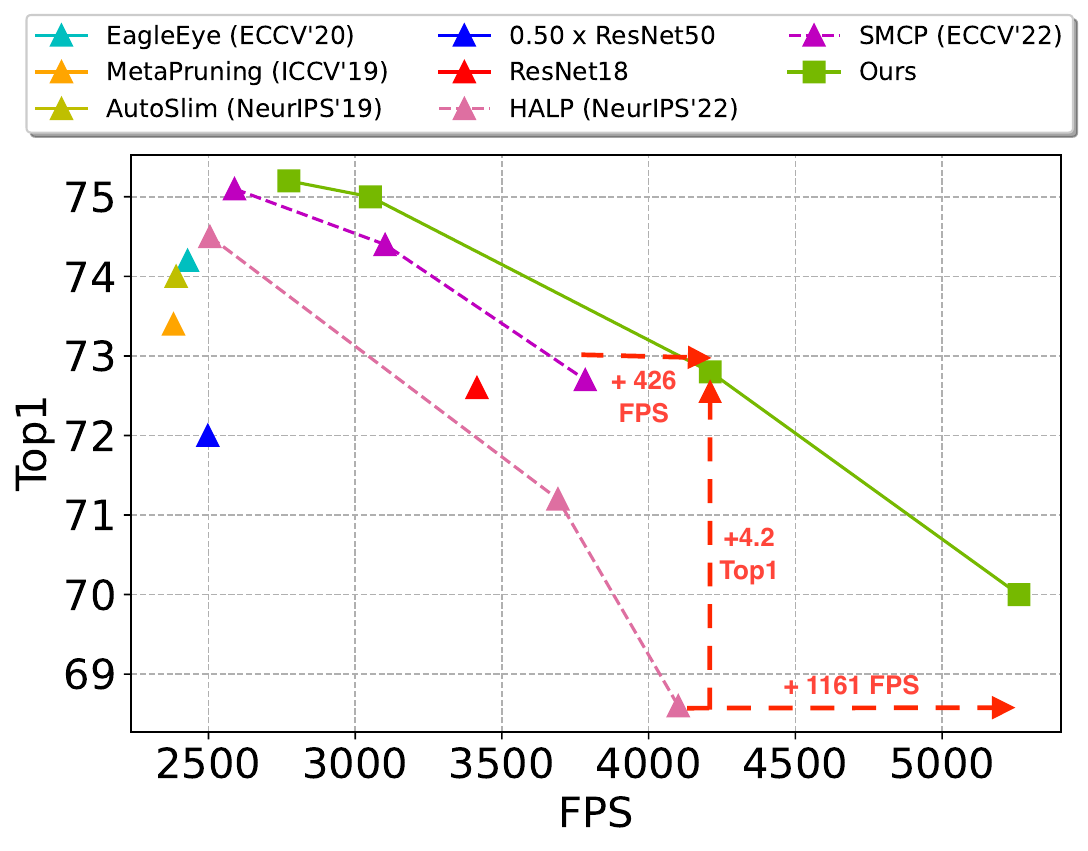}
\end{center}
\end{subfigure}
\hfill
\begin{subfigure}[b]{.5\textwidth}
\begin{center}
   \includegraphics[width=.9\linewidth]{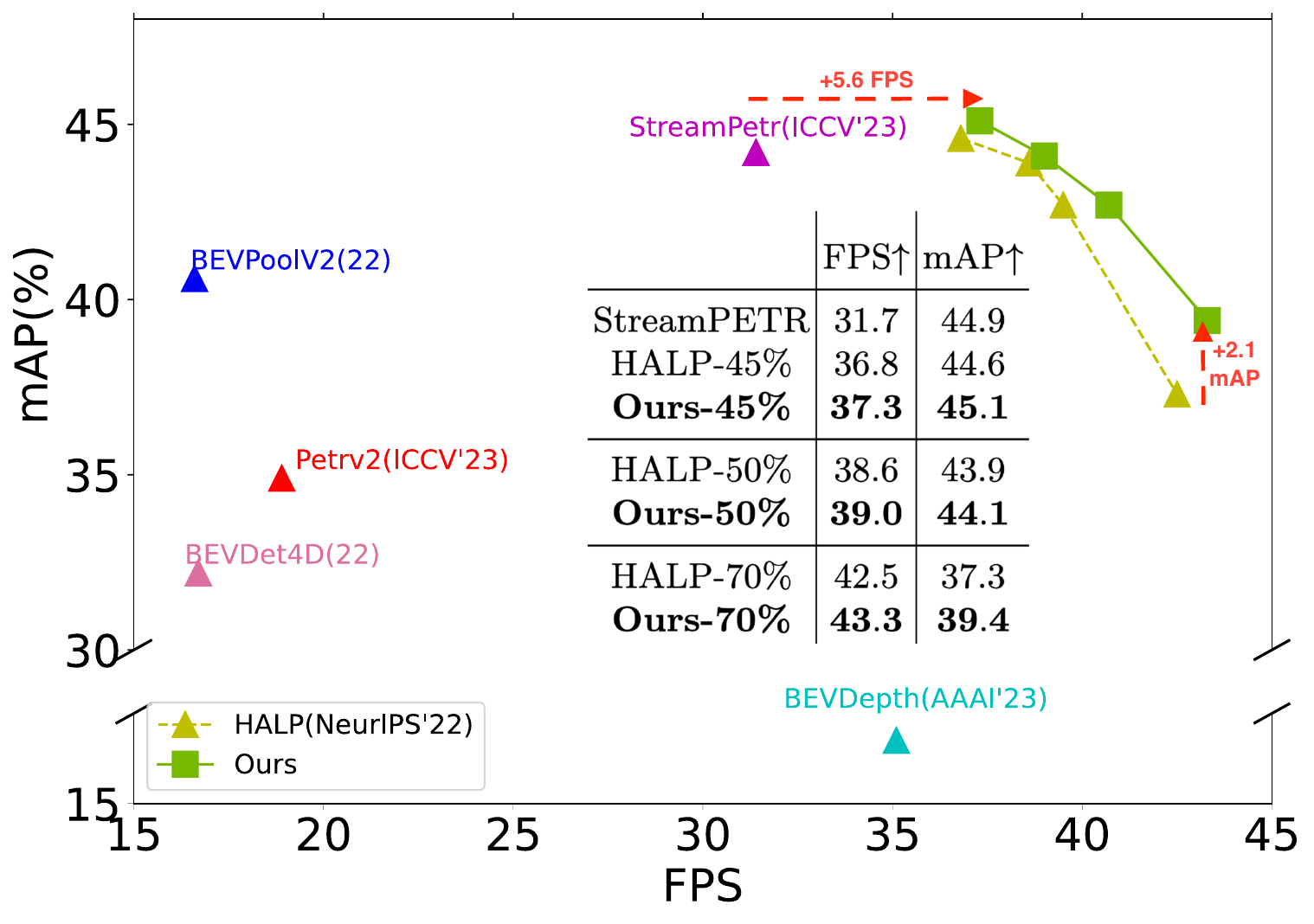}
\end{center}
\end{subfigure}
\caption{\method{} exhibits Pareto dominance across different tasks. In contrast to existing methods: [Left] On Imagenet classification, we achieve a \textbf{6.2\%} relative accuracy gain with a \textbf{2.6\%} FPS speedup, and even greater gains at higher pruning ratio: a \textbf{2\%} relative gain with a substantial \textbf{28.3\%} FPS speedup. [Right] On NuScenes 3D object detection, we observe a \textbf{5.6\%} relative mAP improvement alongside a \textbf{1.8\%} FPS increase.}
\label{fig:teaser}
\vspace{-20pt}
\end{figure}

\begin{abstract}
As we push the boundaries of performance in various vision tasks, the models grow in size correspondingly. To keep up with this growth, we need very aggressive pruning techniques for efficient inference and deployment on edge devices. Existing pruning approaches are limited to channel pruning and struggle with aggressive parameter reductions. In this paper, we propose a novel multi-dimensional pruning framework that jointly optimizes pruning across channels, layers, and blocks while adhering to latency constraints. We develop a latency modeling technique that accurately captures model-wide latency variations during pruning, which is crucial for achieving an optimal latency-accuracy trade-offs at high pruning ratio. We reformulate pruning as a Mixed-Integer Nonlinear Program (MINLP) to efficiently determine the optimal pruned structure with only a single pass. Our extensive results demonstrate substantial improvements over previous methods, particularly at large pruning ratios. In classification, our method significantly outperforms prior art HALP with a Top-1 accuracy of $\mathbf{70.0}$(v.s. $68.6$) and an FPS of $\mathbf{5262}$ im/s(v.s. $4101$ im/s). In 3D object detection, we establish a new state-of-the-art by pruning StreamPETR~\cite{wang2023exploring} at a $45\%$ pruning ratio, achieving higher FPS ($\mathbf{37.3}$ vs. $31.7$) and mAP ($\mathbf{0.451}$ vs. $0.449$) than the dense baseline. 

. 
\end{abstract}
\section{Introduction}

Deep neural networks have become the de-facto standards of advanced computer vision applications, ranging from image classification~\cite{he2016deep} to object detection~\cite{liu2016ssd} and segmentation~\cite{long2015fully}. Contemporary networks~\cite{yang2022moat, dai2021coatnet, wang2023exploring} usually consist of both convolutional neural network (CNN) based feature extractors and transformer blocks to capture global cues. As the performance advances, the models swell in size correspondingly, containing millions or even billions of parameters~\cite{kirillov2023segment}. This growth in model size presents challenges for deployment on resource-constrained edge devices, hinders real-time inference tasks such as autonomous driving, and incurs significant costs for training and inference on cloud systems. Pruning~\cite{han2015deep, molchanov2019importance, shen2021halp}, which involves removing redundant parameters from the network, has emerged as an effective strategy to reduce the model computation and size to reach real-time requirements without significantly compromising its accuracy. To keep pace with the ever-expanding model sizes, we need \textit{very aggressive pruning} techniques to significantly reduce latency for efficient and real-time model deployment.


Channel pruning~\cite{li2017pruning,shen2021halp,humble2022soft,molchanov2019importance, li2020eagleeye, wang2021neural, shen2021when, shen2023hardware}, in particular, has garnered significant attention as an effective pruning technique to reduce model computation, usually 30\% - 50\%, practically without requiring changes in the hardware. Channel pruning involves removing redundant convolution filters identified by some importance criterion~\cite{molchanov2019importance, li2017pruning, lin2020hrank}, usually starting from a pre-trained model. Despite advancements, these methods have two critical limitations. First, channel pruning methods are confined to pruning on the channel level, while we can not avoid the structural removal of entire blocks or layers to achieve the larger pruning ratios required (70\%-90\%). Only a few works~\cite{xu2020layer, wu2023block, tang2023sr, wang2019dbp, elkerdawy2020filter, chen2018shallowing} address layer or block pruning. These methods can provide greater acceleration than channel pruning, but they are restricted to pruning at the layer or block granularity and cannot simultaneously introduce channel sparsity, resulting in suboptimal accuracy.

Second, current pruning approaches to directly reduce inference latency use latency models that only account for variations in output channel count at each layer, ignoring the simultaneous impact of pruning on input channels~\cite{shen2021halp, humble2022soft, castro2022probing, chen2022lap, kong2022spvit, yang2023global, molchanov2022lana, shen2023hardware}. This inaccurate latency estimation leads to sub-optimal trade-offs between accuracy and latency, especially at the larger pruning ratios required for inference on the edge. With large pruning ratios, guiding pruning toward an optimal structure becomes more challenging while adhering closely to the desired latency without precise modeling.

This paper presents a novel pruning framework that effectively overcomes the limitations of existing methods. Specifically, we do not model channels or layers separately. Instead, we first group channels and layers within the same block in our formulation, allowing them to be handled jointly in the optimization process. This unified approach seamlessly integrates channel, layer, and block pruning, enabling us to identify the optimal pruned structure at all levels efficiently. Second, for accurate modeling of latency in different configurations at each layer, we propose the concept of \textit{bilayer configuration latency}, which considers simultaneous variations in both input and output channel counts across all layers. To incorporate these two strategies, we reformulate pruning as a Mixed-Integer Nonlinear Program (MINLP)~\cite{lee2011mixed, burer2012non, bussieck2003mixed}. This allows us to directly solve for the optimal pruned structure adhering to a specific latency budget with only a single pass. As a result, our framework enhances pruning performance with significant latency reductions. All together, we refer to our method as \textbf{M}ulti-\textbf{D}imensional \textbf{P}runing (\method{}). \textit{Our code will be provided upon acceptance.}
\begin{figure}[t!]
    \centering
    \includegraphics[width=\linewidth]{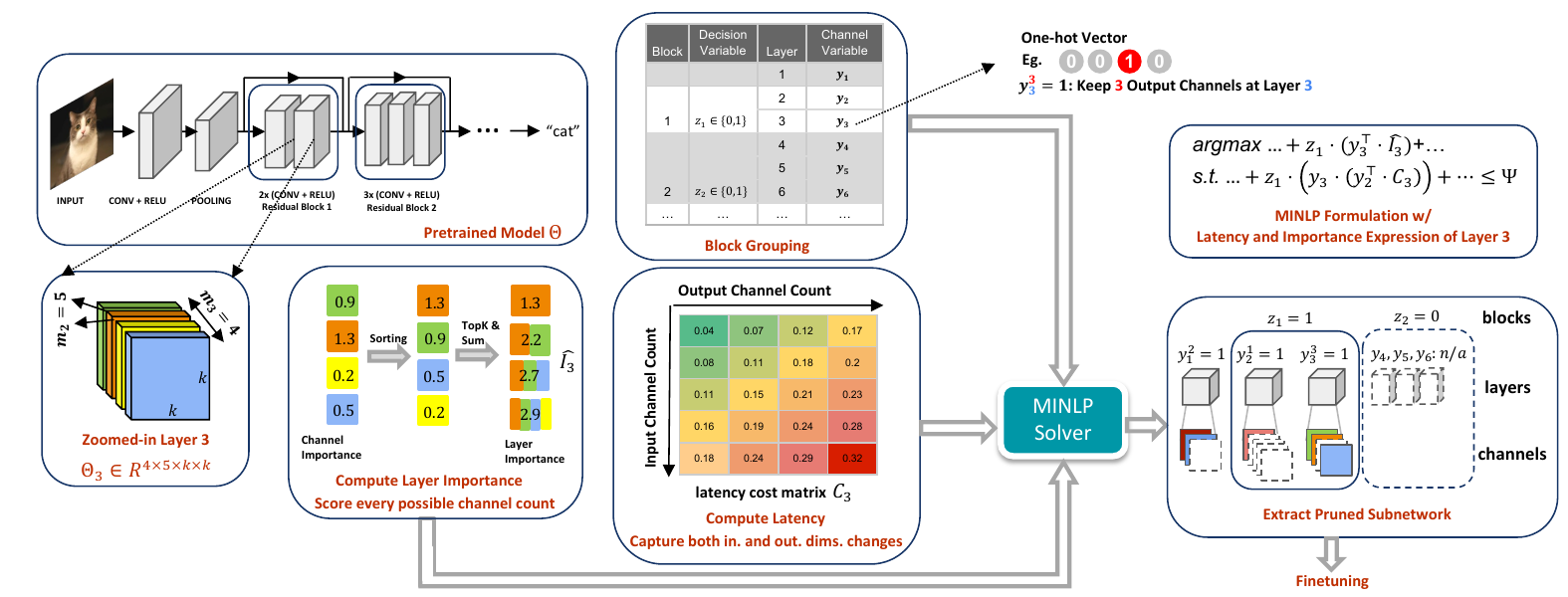}
    \caption{Paradigm of our proposed method \method{}. We start by computing layer importance and constructing latency cost matrices for each layer. We then group layers within the same block and solve an MINLP to optimize pruning decisions at both channel and block levels. Finally, we extract the pruned subnetwork and finetune it.}
    \label{fig:paradigm}
    \vspace{-15pt}
\end{figure}

Our extensive experiments, with a glimpse showcased in Figure~\ref{fig:teaser}, validate our method's superior performance, particularly \textit{at high pruning ratios}. In classification with an aggressive $85\%$ pruning ratio, we significantly outperform the previous work, HALP~\cite{shen2021halp}, with a speed increase of +$\mathbf{1161}$ im/s and an accuracy improvement of +$\mathbf{1.4}$. For 3D object detection, we prune the StreamPETR ~\cite{wang2023exploring} model comprising of a CNN feature extractor and transformer based decoder. We establish a new state-of-the-art at a $45\%$ pruning ratio, achieving higher FPS ($\mathbf{37.3}$ vs. $31.7$) and mAP ($\mathbf{0.451}$ vs. $0.449$) than the dense baseline. At a larger pruning ratio of $70\%$, we significantly outpace HALP~\cite{shen2021halp} in both FPS ($\mathbf{43.3}$ vs. 42.5) and mAP ($\mathbf{0.394}$ vs. 0.373).
We summarize our contributions as follows:
\begin{itemize}
    \item We introduce a \textit{block grouping} strategy for simultaneous channel and block pruning, allowing collective decision-making in optimization.
    \item We propose a method to accurately formulate latency for different layer configurations, capturing variations in both input and output channels.
    \item We organically combine the above strategies with a novel pruning framework redefining pruning as a Mixed-Integer Nonlinear Program (MINLP) which directly solves a globally optimal pruned structure within specific latency constraints efficiently with a single-pass.
    \item We conduct extensive experiments and observe state-of-the-art accuracy-latency trade-offs in a wide range of settings, covering (a) ImageNet~\cite{deng2009imagenet} for classification, Pascal VOC~\cite{everingham2010pascal} for 2D detection, and Nuscenes~\cite{caesar2020nuscenes} for 3D detection (b) with three model architectures: ResNet-50~\cite{he2016deep}, SSD~\cite{liu2016ssd}, and StreamPETR~\cite{wang2023exploring} (c) across various latency reduction pruning ratios.
\end{itemize}

\section{Related Works}
\label{sec:related}
Our work can be categorized as a pruning method in general. We will now provide a brief overview of the field and highlight our differences from the previous approaches. Pruning methods~\cite{lecun1990optimal, hassibi1992second, han2015deep, molchanov2017variational, sun2022disparse, alvarez2016learning, li2017pruning, shen2021halp, molchanov2019importance, lin2020hrank} mostly design importance criterion to rank parameters and remove the lowest-ranked ones, followed by an additional finetuning for accuracy recovery. 

\noindent\textbf{Channel Pruning}
Some pruning methods~\cite{li2017pruning, chin2020towards, he2020learning, he2018soft, yang2018netadapt, he2019filter, lin2020hrank, molchanov2019importance, sun2024towards, shen2021halp, humble2022soft} operate under structural constraints, for example removing convolutional channels\cite{li2017pruning} from CNNs, thus enjoy immediate performance improvement without specialized hardware or library support. Exemplary channel importance criterion relied on metrics like weight norm~\cite{li2017pruning, chin2020towards, he2020learning, he2018soft, yang2018netadapt}, Taylor expansion~\cite{lin2018accelerating, molchanov2019importance, you2019gate}, geometric median~\cite{he2019filter}, and feature maps rank~\cite{lin2020hrank}. Our method leverages the Taylor~\cite{molchanov2019importance} channel importance criterion but extend it to evaluate the configurations of entire layers and blocks, going beyond just pruning channel but also combining layer and block removals. 
  
\noindent\textbf{Layer and Block Pruning}
Channel pruning methods have been effective in reducing performance loss while removing a moderate number of parameters. However, their effectiveness is limited when it comes to more extensive pruning. This is because they focus only on removing channels, but to achieve optimal results with significant pruning, it becomes necessary to remove entire layers or blocks. Only a limited number of works~\cite{chen2018shallowing, elkerdawy2020filter, wang2019dbp, tang2023sr, wu2023block, xu2020layer} focus on pruning layers and blocks. \cite{chen2015compressing} and \cite{elkerdawy2020filter} employ intermediate features at each layer to compute a layer ranking score with linear classifier probes\cite{chen2015compressing} or imprinting\cite{elkerdawy2020filter} and remove the lowest ranked layers. \cite{wang2019dbp} introduces linear classifier probe after each block to check performance and remove blocks with the smallest improvements from the previous. Recent method~\cite{wu2023block} also studies individually removing each block and calculating importance based on the performance drop. 

Though shown to provide larger speedups than channel pruning, all of these approaches~\cite{chen2018shallowing, elkerdawy2020filter, wang2019dbp, tang2023sr, wu2023block, xu2020layer} only operate at the layer or block granularity and fail if we want to introduce channel sparsity simultaneously.  Additionally, some add extra module or parameters like linear probes~\cite{wang2019dbp, chen2015compressing} that require additional training and complicate the process. Our method \textit{seamlessly unites channel and block pruning, which allows us to efficiently determine an optimal pruned structure at both the channel and block levels with just a single pass on the pre-trained model, without any need for extra training or parameters. }

\noindent\textbf{Hardware-aware Pruning} 
Since parameter compression ratio does not directly translate into computation reduction ratio, some works~\cite{li2020eagleeye, wu2020constraint, yang2018netadapt} focus primarily on reducing model FLOPs. Latest methods go one step further and perform hardware-aware pruning which aims to directly reduce the hardware inference latency. The representative work HALP~\cite{shen2021halp} first prepares a latency lookup table for all configurations of prunable parameters measured on the target hardware then formulate pruning as a knapsack problem~\cite{sinha1979multiple}, maximizing total parameter importance while constraining the total associated latency under a given budget. ~\cite{shen2023hardware} later demonstrates that HALP can be applied to autonomous systems to achieve real-time detection performance. To enhance the learning capacity of pruned models, SMCP~\cite{humble2022soft} introduces soft masking within the HALP framework, enabling the reconsideration of earlier pruned weights in an iterative pruning setting. 

Although these methods~\cite{shen2021halp, shen2023hardware, humble2022soft} have made notable progress in accuracy and speed, their reliance on an inaccurate latency estimation leads to suboptimal accuracy-latency trade-offs. They account for changes in output channels but overlook simultaneous variations in input channels caused by pruning the preceding layer. This issue is more pronounced when aiming for large latency reductions, as it becomes more challenging to guide pruning to meet the desired latency budget without accurate latency modeling. In our work, we also focus on hardware-aware pruning \textit{but introduce a more accurate latency modeling technique that accounts for simultaneous variations in both input and output channel counts across all layers, allowing us to determine optimal configurations globally.}

\noindent\textbf{Mixed-Integer Nonlinear Program (MINLP)~\cite{lee2011mixed, burer2012non, bussieck2003mixed}} 
As our strategies of block pruning and accurate latency modeling are unified with a MINLP formulation, we will briefly introduce the field. Formally defined in ~\cite{lee2011mixed}, MINLPs are optimization problems with both integer and continuous variables, where the objective and constraints are nonlinear functions. Effectively and efficiently solving MINLP~\cite{bonami2008algorithmic, d2013mixed, gunluk2010perspective, duran1986outer, fletcher1994solving, bonami2009feasibility, bernal2020improving} is a critical area in optimization research, with key component often involving decomposing the problem into Mixed-Integer Linear Programs (MILP) and Nonlinear Programs (NLP). Recently, conveniently modeling and solving MINLPs in Python has been made possible with Pyomo~\cite{bynum2021pyomo} and MindtPy~\cite{bernal2018mixed}.  In this paper, we use the Outer Approximation (OA)~\cite{duran1986outer,fletcher1994solving} method with Pyomo~\cite{bynum2021pyomo} and MindtPy~\cite{bernal2018mixed} to solve our pruning MINLP.
\section{Methodology}
We will now present our pruning framework. We begin by
establishing preliminaries, defining our goals and declaring relevant notations. We then describe our proposed pruning formulation in detail.

\noindent\textbf{Preliminaries} 
For the neural network with $L$ convolution layers in total, we represent the convolution parameters as $\Theta = \bigcup_{l=1}^{L} \Theta_{l},\Hquad \textrm{s.t.}  \Theta_{l} \in \mathbb{R}^{m_{l} \times m_{l-1} \times K_{l} \times K_{l}}$, where $m_l, m_{l-1}, K_l$ denote the number of output channels, input channels, and kernel size at layer $l$ respectively. Following ~\cite{he2016deep, luo2020neural}, a block is defined as the set of layers(e.g. bottleneck~\cite{he2016deep}) skipped by a residual connection. Suppose there is a total of $B$ blocks in the network $\Theta$. Given a desired inference latency, $\Psi$, our goal is to find the most performant subnetwork $\hat{\Theta} \subseteq \Theta$ through pruning, such that the inference latency of $\hat{\Theta}$ is below the budget $\Psi$ on the target hardware.

Additionally, we declare the following entities(all \textit{1-indexed}):
\begin{table}[H]
    \centering
    \vspace{-15pt}
    \resizebox{\textwidth}{!}
    {
    \begin{tabular}{c|c|c}
    \toprule
         Name&  Notation&  Explanation\\
    \midrule
         \textit{layer2block}&  $\beta(l) \in [1, B]$&  map layer $l$ to ID of the block it belongs to\\
         \textit{layer channel variable}&  $\Vec{y}_l \in \{0,1\}^{m_l}$, one-hot& $y_l^i = 1$ if layer $l$ keeps $i$ out of $m_l$ channels\\
         \textit{block decision variable} & $z_b \in \{0,1\}, b \in [1, B]$ & $z_b=0$ if the entire $b$th block is pruned\\
         \bottomrule
    \end{tabular}
    }
    \vspace{-20pt}
\end{table}

To elaborate, the layer channel variables $\Vec{y}_l$ is defined as one-hot vector, where the index of the hot bit represents \textit{the total number} of selected channels in the pruned network $\hat{\Theta}$, ranging from $1$ to $m_l$.  If the $b$th block is pruned (i.e. $z_b=0$), all layers in this block($\beta(l)=b$) are removed, regardless of the value of $y_l$. In this case, the channel count becomes $0$.

The layer channel variables $\Vec{y}$ and block decision variables $\Vec{z}$ describe the pruning decisions and encode the pruned subnetwork $\hat{\Theta}$, and they are our targets to jointly optimize. In the following sections, we will describe how to solve them based on information we collected from the pre-trained network $\Theta$.
 
\subsection{Multi-Dimensional Pruning (\method{})}
\label{subsec:ours}
To understand how performant a pruned subnetwork $\hat{\Theta}$ is, following previous works~\cite{molchanov2019importance, li2017pruning, lin2020hrank, shen2021halp, humble2022soft}, we leverage importance score as a proxy. The optimal subnetwork $\hat{\Theta}$ is considered to be the one that maximizes the importance score while closely adhering to the latency constraint $\Psi$. With large latency reduction from the original model $\Theta$, we need to consider potential removal of layers and blocks while guiding the pruning decisions with accurate latency estimations. To identify this optimal subnetwork encoded by layer channel variables ($\Vec{y}$) and block decision variables ($\Vec{z}$), we begin by defining two key components, the importance score for different values of $\Vec{y}$ and $\Vec{z}$ and accurate latency estimation for these varying configurations, for each individual layer. The final objective is the aggregation of these components across all layers of the model.


Next, to seamlessly combine complete layer and block removal with channel sparsity, we perform a \textit{block grouping} step which groups the latency and importance expression for all layers within the same block under a single block decision variable. We then formulate the above as a \textit{Mixed-Integer Nonlinear Programming (MINLP)} to jointly determine $\Vec{y}$ and $\Vec{z}$ for an optimal pruned subnetwork $\hat{\Theta}$ at both the channel and block levels. Finally, we extract the pruned subnetwork structure $\hat{\Theta}$ from the solver's output of $\Vec{y}$ and $\Vec{z}$ and carry out a finetuning session on $\hat{\Theta}$ for $E$ epochs to recover the accuracy. All together, we refer to our method as \textbf{M}ulti-\textbf{D}imensional \textbf{P}runing (\method{}). A paradigm diagram of \method{} is demonstrated in Fig.~\ref{fig:paradigm}. We are now going to describe the details of each step.

\noindent \textbf{Layer Importance Computation} As discussed in Sec.\ref{sec:related}, the quality of the pruned subnetworks could be conveniently assessed with channel importance scores. With Taylor importance score~\cite{molchanov2019importance}, 
the importance of the $j$th channel at layer $l$ can be computed as:
\begin{align}
\label{eqn:importance}
    \mathcal{I}_l^j = |g_{\gamma_l^j}\gamma_l^j + g_{\beta_l^j}\beta_l^j|,
\end{align}
where $\gamma$ and $\beta$ are BatchNorm layer's weight and bias for the corresponding channel, and $g_\gamma$ and $g_\beta$ are the gradients.


As the number of channels kept in a layer $l$ is directly encoded by the one-hot variables $\Vec{y}_l$, we associate an importance score for each possible configuration $\Vec{y}_l$ could take, with the one-hot bit index ranging from $1$ to $m_l$. We leverage a greedy approach to derive this. For example, if at layer $l$ pruning keeps $i$ channels (i.e. $y_l^i = 1$), we would like these $i$ channels to be the top-$i$ most important. Therefore, we first rank individual channel importance scores $\Vec{\mathcal{I}}_l$ at layer $l$ in ascending order. Then to compute layer importance score $\hat{I}_l^i$ corresponding to $\Vec{y}_l$ with $y_l^i = 1$, we aggregate the $i$ highest channel importance scores.

Formally, this is expressed as:
\begin{align}
    \hat{\mathcal{I}_l^i} = \sum \text{Top-i}(\Vec{\mathcal{I}}_l), \forall i \in [1, m_l]
    \label{eqn:config_imp}
\end{align}
The vector $\Vec{\hat{\mathcal{I}_l}} \in \mathbb{R}^{m_l}$ fully describes the importance scores for all possible number of channels layer $l$ could take from $1$ to $m_l$. We thereby define the \textit{importance at layer $l$} for a specific configuration $\Vec{y}_l$ as a simple dot-product: $\Vec{y_l}^\top\cdot\Vec{\hat{\mathcal{I}_l}}$. 

\noindent \textbf{Latency Modeling} In order to accurately guide our pruning, we fully describe the latency variations with respect to both the number of its output and input channels and construct a latency cost matrix $\mathbf{C}_l$ for each convolution layer $l$ as follows:
\begin{align}
\label{eqn:latv2}
{
  \mathbf{C}_l =
  \left[ {\begin{array}{cccc}
    T_l(1,1) & T_l(1,2)& \cdots & T_l(1,m_l)\\
    T_l(2, 1) & T_l(2,2) & \cdots & T_l(2, m_l)\\
    \vdots & \vdots & \ddots & \vdots\\
    T_l(m_{l-1}, 1) & T_l(m_{l-1}, 2) & \cdots & T_l(m_{l-1}, m_l)\\
  \end{array} } \right]
}
\end{align}
Here, $T_l$ is a pre-built latency lookup table, which could be measured on the target hardware beforehand as prior works~\cite{shen2021halp, shen2023hardware, humble2022soft}, and $T_l(i, j)$ returns the latency of layer $l$ with $i$ input channels and $j$ output channels, upper-bounded by the total channel count $m_{l-1}$ and $m_l$ in $\Theta$. $\mathbf{C}_l$ enumerates latency corresponding to all possibilities layer $l$ could be, varying the input and output channel numbers. With these configurations encoded in the one-hot layer channel variables $\Vec{y}_{l-1}$ and $\Vec{y}_{l}$, we define the \textit{bilayer configuration latency} at layer $l$ for specific $\Vec{y}_{l-1}$ and $\Vec{y}_{l}$ simply as two dot-products: $\Vec{y_{l}} \cdot (\Vec{y_{l-1}}^\top \cdot \mathbf{C}_l)$.

We can observe that each $\Vec{y}_{l}$ appears twice in the expressions, once at computing latency for layer $l$, and once at layer $l+1$. While this poses some challenges in optimization, it manages to accurately capture the full overall latency landscape of network. This approach enables us to guide the pruning process with more precise latency estimations, significantly improving the precision from previous methods~\cite{shen2021halp, humble2022soft, shen2023hardware} that did not consider the simultaneous contributions from both output and input channel dimension.

\noindent \textbf{Block Grouping} Notice that we define the layer channel variables $\Vec{y}_l$ to only describe channel count from $1$ to $m_l$, excluding the case when pruning removes all channels from layer $l$(i.e. channel count of $0$). This means if we only use variables $\Vec{y}$ to represent the pruned model $\hat{\Theta}$, we cannot represent completely removing a layer from the network; at best, we can reduce it to just one channel, similar to previous methods~\cite{shen2021halp, humble2022soft, shen2023hardware}. This is intentional because arbitrarily pruning a single layer could easily lead to network disconnection, causing discontinuity in the information flow of the network. However, residual blocks are inherently resilient to removal of all their internal layers at once, as the skip connection allows information to bypass the removed layers, preserving gradient flow. 

To handle the removal of an entire residual block structure, we introduce block grouping where layers are grouped into the block it belongs to. Specifically, we parse the network architecture to obtain the \textit{layer2block} mapping $\beta(l)$ for every layer $l$. Then we group all importance and latency expressions within the same block under a single block decision variable. If pruning decides to remove the $b$th block, the importance and latency contributions from all layers within that block, where $\beta(l) = b$, should be simultaneously set to $0$.

We model this group decision with the binary block decision variables $z_b$. Subsequently, for each layer $l$, we first determine whether its associated block decision variable, denoted by $z_{\beta(l)}$, is active ($z_{\beta(l)}=1$). Only if it is active, we evaluate the layer importance and latency expressions determined by $\Vec{y}$; otherwise, they are simply zeroed. The \textit{importance} for layer $l$ is determined by both $y_l$ and $z_{\beta(l)}$, and can be expressed as $z_{\beta(l)} \cdot( \Vec{y_l}^\top\cdot\Vec{\hat{\mathcal{I}_l}})$. Similarly, the \textit{bilayer configuration latency} at convolution layer $l$ can be represented as $z_{\beta(l)} \cdot (\Vec{y_{l}} \cdot (\Vec{y_{l-1}}^\top \cdot \mathbf{C}_l))$.

Block removal is properly handled by giving the block decision variables $\Vec{z}$s higher 'priority' than the layer channel variables $\Vec{y}$s. For example, deactivating $z_1$ results in the exclusion of all layers within the first block(where $\beta(l) = 1$) by simultaneously setting their importance and latency expressions to $0$, regardless of the values taken by their $\Vec{y}_l$s. Also, notice that for the \textit{layers that do not belong to any block structures, their corresponding $\Vec{z}$s are simply always $1$.}

\noindent \textbf{Solve MINLP} We aim to jointly determine the optimal layer channel and block decisions($\Vec{y}$ and $\Vec{z}$) that maximize the summation of their \textit{importance scores} while ensuring the cumulative \textit{bilayer configuration latency} remains below the budget $\Psi$. Formally, this can be represented with the following Mixed-Integer Nonlinear Programming (MINLP) formulation:
\begin{align}   
\label{eqn:program}
\argmax_{\Vec{y}, \Vec{z}} & \quad \sum_{l=1}^L  z_{\beta(l)} \cdot( \Vec{y_l}^\top\cdot\Vec{\hat{\mathcal{I}_l}}) \\
\textsc{s.t.} \quad \sum_{l=1}^L  z_{\beta(l)} \cdot & (\Vec{y_{l}} \cdot (\Vec{y_{l-1}}^\top \cdot \mathbf{C}_l)) \leq \Psi \nonumber
\end{align}

We restrict all decision variables $\Vec{y}$ and $\Vec{z}$ to \textbf{\textit{binary}} values, while the layer importances $\Vec{\hat{\mathcal{I}}}_l$ and latency cost matrices $\mathbf{C}_l$ contain \textbf{\textit{floating-point}} numbers, hence making the program mixed-integer in nature. Recall each layer channel variable $\Vec{y}_l$ is one-hot, which can be formally formulated as follows as an additional constrain to Eqn.~\ref{eqn:program}:

\begin{align}
    \Vec{y_l}^\top &\cdot \mathbf{1} = 1, \Hquad\forall l\in [1, L]
\end{align}

To solve this MINLP~\ref{eqn:program}, we leverage the Python numerical decomposition framework Pyomo~\cite{bynum2021pyomo} and MindtPy~\cite{bernal2018mixed}, and employ the Feasibility Pump (FP) method~\cite{bonami2009feasibility} to enhance efficiency. Since we jointly optimize all variables, we can directly determine a globally optimal set of $\Vec{y}$ and $\Vec{z}$ \textbf{\textit{with only a single pass. }}

\begin{algorithm}[t]
    \caption{\method{} Framework}\label{euclid}
    \label{algo:1}
    \textbf{Input:} Pretrained weights $\Theta$, latency lookup table $T$, total finetuning epochs $E$, training dataset $\mathcal{D}$, latency budget $\Psi$
    \begin{algorithmic}[1]
    \State{Declare layer channel variables $\Vec{y}$ and block decision variables $\Vec{z}$}
        \State{\texttt{//Layer Importance Computation}}
        \For{sample $(x, y)$ in $\mathcal{D}$}
        \State{Perform forward pass and backward pass with $\Theta$}
        \State{Calculate Taylor channel importance score 
 $\Vec{\mathcal{I}}_l$} (Eqn.~\ref{eqn:importance})
        \State{Calculate and accumulate layer importance score $\Vec{\hat{\mathcal{I}}}_l$ (Eqn.~\ref{eqn:config_imp})}
        \EndFor
        \State{Construct \textit{importance} expression: $\Vec{y}_l^\top \cdot \Vec{\hat{\mathcal{I}}}_l$}
        \State{\texttt{//Latency Modeling}}
        \State{Construct latency matrices $\mathbf{C}_l$ (Eqn.\ref{eqn:latv2})}
        \State{Construct \textit{bilayer configuration latency} expression: $\Vec{y_{l}} \cdot (\Vec{y_{l-1}}^\top \cdot \mathbf{C}_l)$}
        \State{\texttt{//Block Grouping}}
        \State{Obtain the \textit{layer2block} mapping $\beta(l)$ for each layer $l$}
        \State{Group importance and latency expressions under $\Vec{z}$}
        \State{\texttt{//Solve MINLP}}
        \State{Set up the MINLP (Eqn. \ref{eqn:program}) and solve it with Pyomo and MindtPy}
        \State{\texttt{//Extract Pruned Structure}}
        \State{Extract pruned subnetwork $\hat{\Theta}$ from solver output $\Vec{y}$ and $\Vec{z}$}
        \State{Finetune the pruned model $\hat{\Theta}$ as usual for $E$ epochs}
    \end{algorithmic}
    \label{algo:pseudocode}
\end{algorithm}

\noindent \textbf{Extract Pruned Structure} Once we solved the MINLP program~\ref{eqn:program}, we proceed to extract the pruned subnetwork $\hat{\Theta}$ based on the variables $\Vec{y}$ and $\Vec{z}$ determined by the solver. If block decision variable $z_b$ is set to $0$ for a particular block $b$, we completely remove that block in $\hat{\Theta}$ and disregard the solver's output for the layer channel variables($\Vec{y}_l$) of the layers within that block (where $\beta(l) = b$). On the other hand, if the block is active with $z_b = 1$ and the solver returns the value of $\Vec{y}$ with $\Vec{y}_l^i = 1$, we keep $i$ channels in $\hat{\Theta}$ at layer $l$ according to $\text{ArgTopK}(\Vec{\mathcal{I}}_l, i)$, mapping layer importance $\Vec{\hat{\mathcal{I}}}_l$ in Eqn.\ref{eqn:config_imp} back to the $i$ top-performing channels. 

\noindent In practice, since the layer importances $\Vec{\hat{\mathcal{I}}}$ are built from Taylor score~\cite{molchanov2019importance} measured using gradient information from data batches, we perform pruning after one epoch when the model has seen all of the samples in the dataset and accumulate their importance estimation. After pruning, we finetune the pruned model $\hat{\Theta}$ for a duration of $E$ epochs to recover accuracy. An algorithmic description of the above process is provided in Algorithm~\ref{algo:pseudocode}.

\section{Experiments}

To validate the proposed method, we perform extensive experiments across a comprehensive set of scenarios. We demonstrate our pruning results on 3 tasks: image classification with ImageNet~\cite{deng2009imagenet} and ResNet50~\cite{he2016deep}, 2D object detection with Pascal VOC~\cite{everingham2010pascal} and SSD~\cite{liu2016ssd}, and 3D object detection with Nuscenes~\cite{caesar2020nuscenes} and StreamPETR~\cite{wang2023exploring}. 

Our method improves upon previous approaches by providing \textbf{(a)} more accurate latency estimation and \textbf{(b)} the ability to handle the removal of entire layer and block structures. These improvements are reflected in our superior result, as we achieve a new state-of-the-art with a significantly better accuracy-latency trade-off compared to prior arts~\cite{shen2021halp, humble2022soft} and other competitive baselines~\cite{li2020eagleeye, wang2021neural}, especially at higher pruning ratios. To provide a comprehensive understanding of our proposed pruning framework, we also conducted an in-depth ablation study, highlighting the individual contribution of our improvement from \textbf{(a)} and \textbf{(b)}.

\noindent \textbf{Settings} For ResNet50 and SSD, we aimed to optimize their inference latency on the Nvidia TITAN V GPU with batch size of $256$. When pruning StreamPETR~\cite{wang2023exploring}, we adapted our approach to target the Nvidia GeForce RTX 3090 GPU batch size of $1$, aligning with the focus of StreamPETR's original paper. This allowed us to fairly evaluate the speedup benefits and also demonstrate the generalization of our method targeting different hardware platforms.

All of our trainings use 8 Nvidia Tesla V100 GPUs and conducted with PyTorch~\cite{paszke2017automatic} V1.4.0. Our CPU is Intel Xeon E5-2698 and is used to solve the MINLP optimization(Eqn. ~\ref{eqn:program}). 

\begin{table*}[t!]
    \centering
    \vspace{-0.2cm}
    \resizebox{\textwidth}{!}
    {
        \begin{tabular}{lcccc}
            \toprule
            \rowcolor{lgray} 
            \textsc{Method} & \textsc{Top-1 Acc}($\%$)$\uparrow$ & \textsc{Top-5 Acc}($\%$)$\uparrow$  & \textsc{FLOPs}($\times e^9$)$\downarrow$ & \textsc{FPS(im/s)}$\uparrow$ \\
            \midrule
            \multicolumn{5}{c}{\textbf{ResNet50~\cite{he2016deep}}}\\
            \textsc{Dense} & $76.2$&$92.9$ & $4.1$&$1019$\\
            \midrule
            \textsc{ResConv-Prune}\cite{xu2020layer} & $70.0$ & $90.0$ & $1.6$ & $--$ \\
            \textsc{DBP-0.5}\cite{wang2019dbp} & $72.4$ & $--$ & $--$ & $1630^*$ \\
            \textsc{LayerPrune$_7$-Imprint}\cite{elkerdawy2020filter} & $74.3$ & $--$ & $--$ & $1828^*$ \\
            \textsc{MetaPrune}\cite{liu2019metapruning} & $73.4$ & --& $1.0$ & $2381$ \\
            \textsc{AutoSlim}\cite{yu2019autoslim} & $74.0$ & --& $1.0$ & $2390$ \\
            \textsc{GReg-2}\cite{wang2021neural} & $73.9$ & --& $1.3$ & $1514$ \\
            \textsc{HALP-$70\%$}\cite{shen2021halp} & $74.5$ &$91.8$ & $1.2$ & $2597$ \\
            \textsc{SMCP-$70\%$}\cite{humble2022soft} & $\mathbf{74.6}$ &$92.0$ & $1.0$ & $2947$ \\
            \rowcolor{lgreen} \textbf{\textsc{Ours-70\%}} & $\mathbf{74.6}$ &$\mathbf{92.2}$ & $1.1$ & $\mathbf{3092}$ \\
            \midrule
            \textsc{HALP-$85\%$}\cite{shen2021halp} & $68.1$ &$88.4$& $0.6$ & $3971$ \\
            \rowcolor{lgreen} \textbf{\textsc{Ours-85\%}} & $\mathbf{70.0}$ &$\mathbf{89.3}$& $0.5$ & $\mathbf{5306}$ \\
            \midrule
            \midrule
            \multicolumn{5}{c}{\textbf{ResNet50 - EagleEye~\cite{li2020eagleeye}}}\\
            \textsc{Dense}~\cite{li2020eagleeye} & $77.2$ &$93.7$& $4.1$&$1019$\\
            \bottomrule
 \textsc{EagleEye-1G}\cite{li2020eagleeye} & $74.2$ &$91.8$ & $1.0$ & $2429$ \\
 \textsc{HALP-70\%}\cite{shen2021halp} & $74.5$ &$91.9$ & $1.2$ & $2597$ \\
  \textsc{SMCP-70\%}\cite{humble2022soft} & $75.1$ &$92.3$ & $1.1$ & $2589$ \\
    \rowcolor{lgreen}
  \textbf{\textsc{Ours-65\%}} & $\mathbf{75.2}$ &$\mathbf{92.5}$& $1.3$ & $\mathbf{2774}$ \\
  \rowcolor{lgreen}
  \textbf{\textsc{Ours-70\%}} & $75.0$ &$92.2$& $1.2$ & $3052$ \\
  \midrule
  \textsc{HALP-80\%}\cite{shen2021halp} & $71.2$ &$90.1$& $0.7$ & $3691$ \\
  \textsc{SMCP-80\%}\cite{humble2022soft} & $72.7$ &--& --& $3784$ \\
  \rowcolor{lgreen}
  \textbf{\textsc{Ours-80\%}} & $\mathbf{72.8}$ &$\mathbf{90.9}$& $0.7$ & $\mathbf{4210}$ \\
  \midrule
\textsc{HALP-85\%}\cite{shen2021halp} & $68.6$ &$88.5$& $0.6$ & $4101$ \\
\rowcolor{lgreen}
\textbf{\textsc{Ours-85\%}} & $\mathbf{70.0}$ &$\mathbf{89.2}$& $0.5$ & $\mathbf{5262}$ \\
\bottomrule
        \end{tabular}
    }
    \caption{\textbf{ImageNet results with ResNet-50}. FPS is measured on NVIDIA TITAN V with batch size of $256$. Results with similar FPS are grouped. $-X\%$ denote the pruning ratio. $^*$ denotes latency estimated from the reported ratio. Ours achieve much better accuracy-FPS tradeoffs than the baselines, specifically when pruning ratio is large. Averaged results over two runs.}
    \label{table:imageresnet50}
\vspace{-20pt}
\end{table*}
\subsection{Classification Results on ImageNet} In Table~\ref{table:imageresnet50}, We report our pruning results and comparison with the baseline methods on ResNet50~\cite{he2016deep} and ImageNet~\cite{deng2009imagenet}. We evaluate these results using Top-1 Accuracy and Top-5 Accuracy to gauge the recovered accuracy after fine-tuning. In addition, we include inference FPS (im/s, i.e. images per second) to directly showcase the speedups on the target hardware. We also present FLOPs for completeness.

Compared with previous methods like HALP~\cite{shen2021halp} and SMCP~\cite{humble2022soft}, we achieve a significantly improved accuracy-latency trade-off. For instance, SMCP reaches a Top-1 accuracy of $72.7$ with an inference speed of $3784$ im/s; our method slightly surpasses its Top-1 with an accuracy of $\mathbf{72.8}$ but with a considerably faster inference speed of \textbf{4210} im/s. With larger pruning, HALP achieves a Top-1 accuracy of $68.6$ with an inference speed of $4101$ im/s, our method significantly outperforms it with a Top-1 accuracy of $\mathbf{70.0}$ and an impressive FPS of $\mathbf{5262}$ im/s. Notably, we can observe from Table~\ref{table:imageresnet50} that our method particularly excels when \textbf{\textit{targeting high FPS}} with substantial pruning from pre-trained models, corroborating the effectiveness of improvements from our method. Our improvements could be observed more clearly in the FPS v.s. Top-1 Pareto curve displayed in Figure~\ref{fig:teaser}. 

We also include direct comparison with methods~\cite{xu2020layer, wang2019dbp, elkerdawy2020filter} which also specifically discuss layer and block removal. As shown in Table~\ref{table:imageresnet50}, our results are significantly better. For instance, compared to LayerPrune~\cite{elkerdawy2020filter}, we achieve a higher Top-1 accuracy ($\mathbf{74.6}$ vs. $74.3$) and a substantially greater FPS ($\mathbf{3092}$ vs. $1828$).

\subsection{2D Object Detection Results on PascalVOC} 
To illustrate the broad applicability of our approach, we also conducted experiments in the realm of 2D object detection using the widely recognized Pascal VOC dataset~\cite{everingham2010pascal}. In Figure~\ref{fig:pascal}, we present the outcomes of our pruning methodology applied to an SSD512~\cite{liu2016ssd} model with a ResNet50 backbone. Our performance is assessed against various competitive baselines, including HALP~\cite{shen2021halp} and SMCP~\cite{humble2022soft}. We depict the Pareto frontier, showcasing the trade-off between FPS and mean Average Precision (mAP).

Our results distinctly outshine existing methods in the field, marking a substantial advancement. In direct comparison to SMCP, our approach consistently achieves significantly higher mAP scores across various inference FPS levels. For instance, we outperform SMCP with an mAP of $\mathbf{79.2}$ (compared to $78.3$) while also slightly increasing the FPS to $\mathbf{146.4}$ (compared to $144.4$).Notably, our pruned model even surpasses the mAP of the pre-trained dense SSD512-RN50 by a margin($\mathbf{80.0}$ v.s. $78.0$)while substantially enhancing its FPS($\mathbf{125.4}$ v.s. $68.2$).

\begin{figure}[t!]
\begin{subfigure}[b]{.47\textwidth}
\begin{center}
   \includegraphics[width=\linewidth]{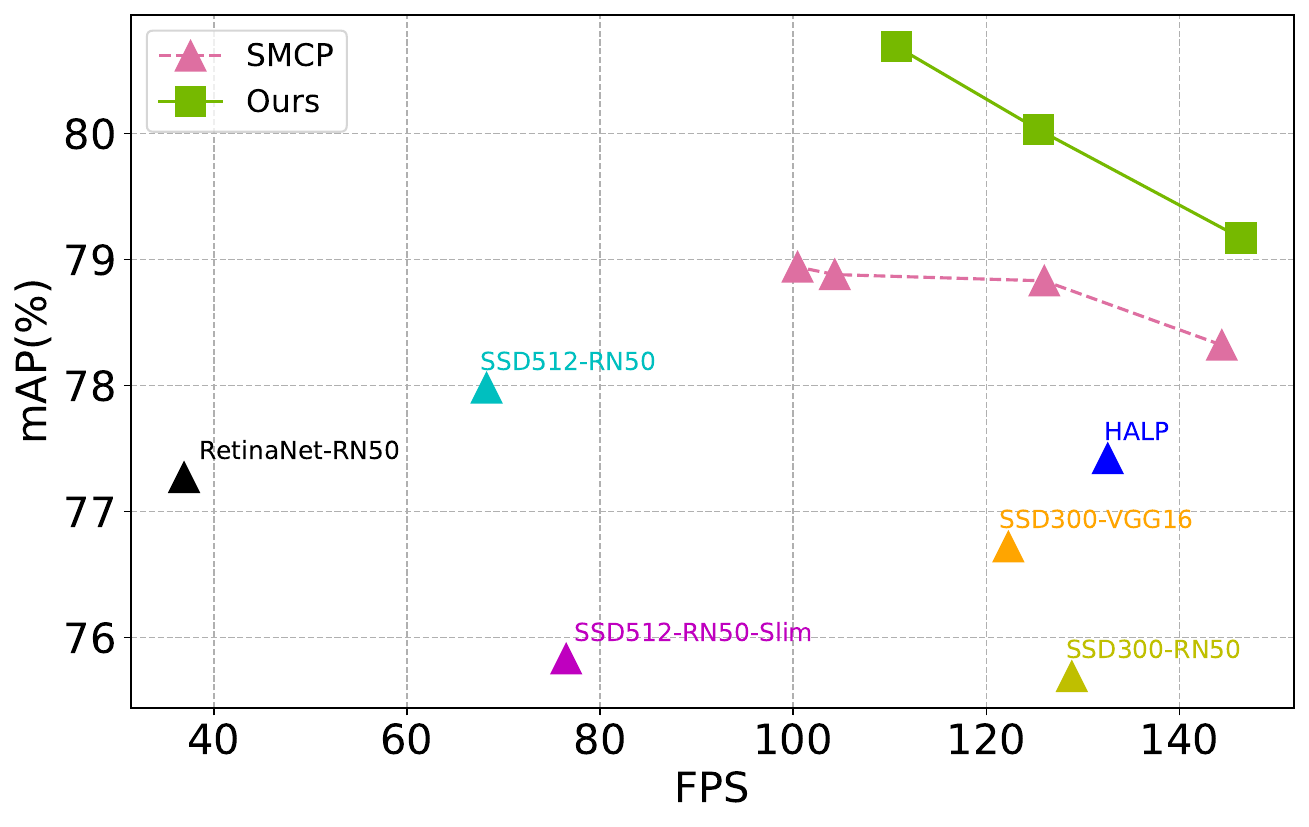}
\end{center}
\caption{FPS versus mAP are plotted(top-right is better). FPS measured on NVIDIA TITANV.}
\label{fig:pascal}
\end{subfigure}%
\hfill
\begin{subfigure}[b]{.47\textwidth}
\begin{center}
   \includegraphics[width=\linewidth]{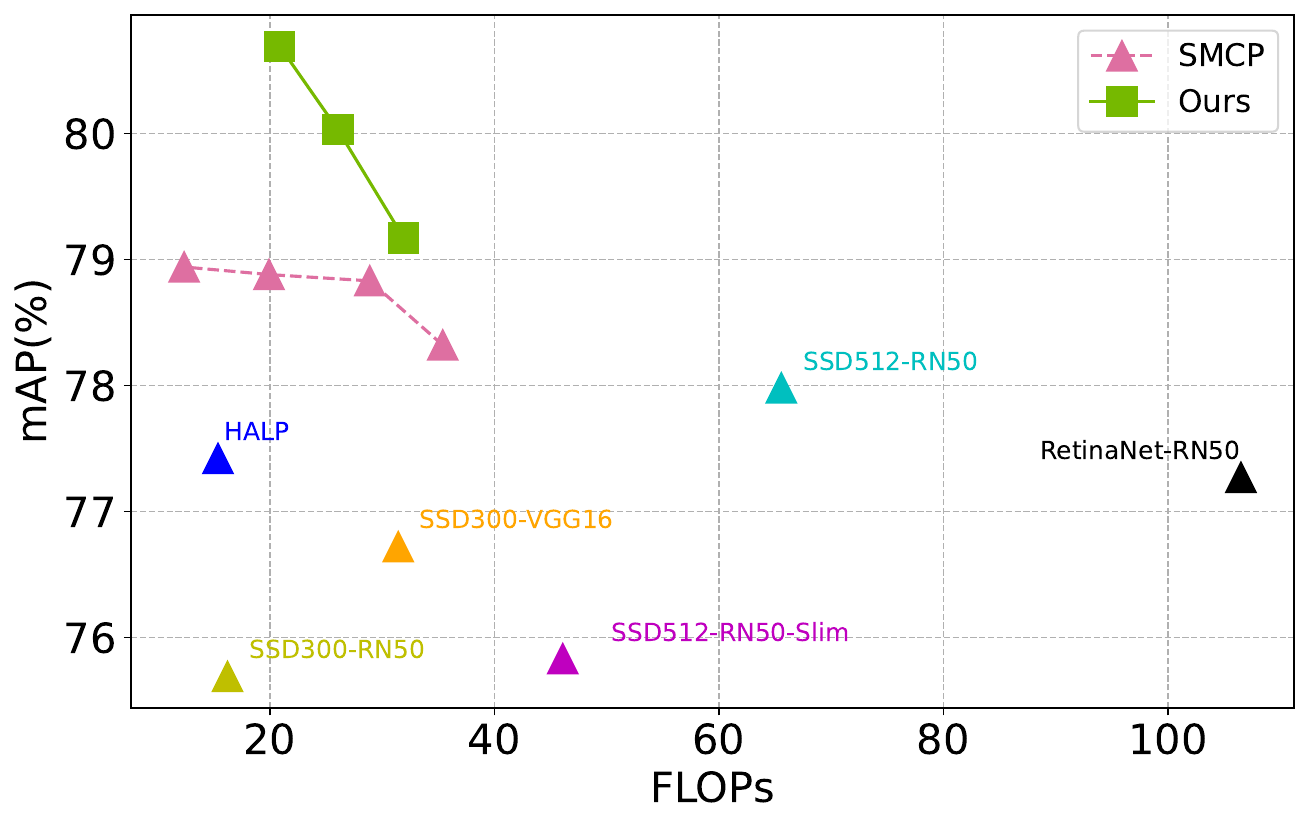}
\end{center}
\caption{FLOPs versus mAP are plotted(top-left is better).}
\label{fig:pascal}
\end{subfigure}
\caption{\textbf{PascalVOC results with SSD512}. FPS is measured on NVIDIA TITAN V with batch size of $256$. Ours achieve much better mAP-FPS and mAP-FLOPs tradeoffs than the baselines.}
\end{figure}

\begin{table*}[t!]
\centering
\begin{tabular}{c|c|c|ccccc|c}
\toprule
\rowcolor{lgray}
\textsc{Method} & \textsc{mAP$\uparrow$} & \textsc{NDS$\uparrow$} & \textsc{mATE$\downarrow$} & \textsc{mASE$\downarrow$} & \textsc{mAOE$\downarrow$} & \textsc{mAVE$\downarrow$} & \textsc{mAAE$\downarrow$} & \textsc{FPS$\uparrow$} \\
\midrule
\textsc{BEVPoolv2}\cite{huang2022bevpoolv2} & $0.406$ & $0.526$ & $0.572$ & $0.275$ & $0.463$ & $0.275$ & $0.188$ & $16.6$ \\
\textsc{BEVDet4D}\cite{huang2022bevdet4d}& $0.322$ & $0.457$ & $0.703$ & $0.278$ & $0.495$ & $0.354$ & $0.206$ & $16.7$ \\
\textsc{PETRv2}\cite{liu2023petrv2} & $0.349$ & $0.456$ & $0.700$ & $0.275$ & $0.580$ & $0.437$ & $0.187$ & $18.9$ \\
\textsc{Sparse4Dv2}\cite{lin2023sparse4d} & $0.439$ & $0.539$ & $0.598$ & $0.270$ & $0.475$ & $0.282$ & $0.179$ & $20.3$ \\
\textsc{StreamPETR}\cite{wang2023exploring} & $0.449$ & $0.546$ & $0.613$ & $0.267$ & $0.413$ & $0.265$ & $0.196$ & $31.7$ \\
\textsc{HALP-45\%}~\cite{shen2021halp} & $0.446$ & $0.547$ & $0.605$ & $0.271$ & $0.409$& $0.269$& $0.211$& $36.8$ \\
\rowcolor{lgreen}
\textbf{\textsc{Ours-45\%}} & $\mathbf{0.451}$ & $\mathbf{0.551}$ & 0.596& 0.272& 0.413& $0.259$& $0.207$& $\mathbf{37.3}$ \\
\midrule
\textsc{HALP-50\%}~\cite{shen2021halp} & $0.439$ & $0.543$ & 0.605 & 0.270 & 0.419 & 0.265 & 0.211 & $38.6$ \\
\rowcolor{lgreen}
\textbf{\textsc{Ours-50\%}} & $\mathbf{0.441}$ & $\mathbf{0.544}$ & 0.606 & 0.269 & 0.421 & 0.268 & 0.205& $\mathbf{39.0}$ \\
\midrule
\textsc{HALP-60\%}~\cite{shen2021halp} & $\mathbf{0.427}$ & $\mathbf{0.533}$ & 0.620 & 0.269 & 0.438 & 0.271 &0.209 &$39.5$ \\
\rowcolor{lgreen}
\textbf{\textsc{Ours-60\%}} & $\mathbf{0.427}$ & $0.532$ & 0.608 &  0.272 & 0.457 &0.269 &0.207 & $\mathbf{40.7}$ \\
\midrule
\textsc{HALP-70\%}~\cite{shen2021halp} & $0.373$ & $0.489$ & 0.674 & 0.277 & 0.534 & 0.293 & 0.197 & $42.5$ \\
\rowcolor{lgreen}
\textbf{\textsc{Ours-70\%}} & $\mathbf{0.394}$ & $\mathbf{0.512}$ & 0.642 & 0.275 & 0.449 & 0.278 & 0.204 & $\mathbf{43.3}$ \\
\bottomrule
\end{tabular}
\caption{\textbf{Nuscenes results with StreamPETR}. FPS is measured on NVIDIA GeForce RTX 3090 with batch size of $1$. Results with similar FPS are grouped. $-X\%$ denote the pruning ratio. Ours achieve much better accuracy-FPS tradeoffs than HALP and even surpass performance of dense StreamPETR with much higher FPS.}
\label{table:nuscenes}
\vspace{-10pt}
\end{table*}
\subsection{3D Object Detection Results on Nuscenes}
So far, we have shown that our pruning method is effective for models composed entirely of convolutional layers, such as ResNet50 and SSD. Modern systems deploy convolutional layers for features extraction and transformer layers for capturing global cues~\cite{yang2022moat, dai2021coatnet}. In this section, we explore our pruning effectiveness for these hybrid models. We focus on the challenging task of 3D object detection, using the widely recognized Nuscenes~\cite{caesar2020nuscenes} dataset and the state-of-the-art model StreamPETR~\cite{wang2023exploring}, composed of a heavy CNN-based encoder and a transformer-based decoder. Our analysis of the system's latency revealed that the CNN-based encoder has a higher latency ($16.7 ms$) than the transformer decoder ($14ms$). This indicates that applying our method to the convolutional layers can still effectively accelerate the entire network.

Detailed results and comparisons with several competitive baselines are presented in Table ~\ref{table:nuscenes}. Our evaluation incorporated a diverse set of metrics commonly adopted for 3D object detection tasks~\cite{caesar2020nuscenes, wang2023exploring, lin2023sparse4d}, including the mean Average Precision (mAP) and Normalized Detection Score (NDS). Additionally, we report the FPS to highlight the improvements in speed.

Significantly, when compared to the dense pre-trained StreamPETR model, our technique achieved a substantial acceleration of approximately $\mathbf{18\%}$, resulting in an impressive $\mathbf{37.3}$ FPS as opposed to the baseline's $31.7$ FPS. Importantly, this speed boost was achieved without sacrificing performance: our pruned model attained superior mAP ($\mathbf{0.451}$ vs. $0.449$) and NDS ($\mathbf{0.551}$ vs. $0.546$). In comparison to the previous pruning method HALP~\cite{shen2021halp}, our approach exhibited remarkable improvements in accuracy-latency trade-offs across various pruning ratios. For instance, HALP managed to produce a pruned StreamPETR model with an mAP of $0.373$, an NDS of $0.489$, and an inference FPS of $42.5$. In contrast, our approach surpassed these results, achieving an mAP of $\mathbf{0.394}$, an NDS of $\mathbf{0.512}$, and an inference FPS of $\mathbf{43.3}$.

\subsection{Ablation Study}
\subsubsection{} 
As discussed in detail in Sec.~\ref{subsec:ours}, our pruning method introduces two key improvements from prior methods: \textbf{(a)}``\textit{bilayer configuration latency}" for accurate latency modeling; \textbf{(b)} "block grouping" for integration of block removal with channel sparsity.

We'll now explore the individual impacts of \textbf{(a)} and \textbf{(b)} on pruning performance. The \textit{baseline} here is thus a bare latency pruning algorithm without both \textbf{(a) and (b)}. We then ablate each component by separately adding them on top of the baseline to check inidividual improvement. The baseline performance is depicted in Fig.~\ref{fig:ablation} with label ``\textit{Baseline}".


\noindent\textbf{Bilayer Configuration Latency}
In this setting, we add our ``\textit{bilayer configuration latency}" on top of the baseline but drop the \textit{block grouping} step to exclude the block decision variables from the MINLP program ~\ref{eqn:program}. This variant accurately estimates the latency impacts of pruning by considering variations in both input and output channel counts, but it cannot handle removal of entire blocks. 
The result, labeled ``\textit{Ours(Only Bilayer Latency)}" in Fig.~\ref{fig:ablation}, show a markedly better accuracy-latency tradeoff than the baseline, demonstrating its effectiveness even when used alone.

\noindent\textbf{Block Grouping}
In this setting, we add our ``\textit{block grouping}" step to the baseline but do not use our ``\textit{bilayer configuration latency}" to model latency impacts from pruning. Instead, we use previous methods' latency modeling~\cite{shen2021halp, humble2022soft, shen2023hardware}, which only account for variations in output channel counts. This variant can effectively handle the removal of block strucutres to accommodate high pruning ratios but but cannot accurately estimate latency impacts by considering changes in both input and output channel counts. The results, labeled ``\textit{Ours(Only Block Grouping)}" in Fig.~\ref{fig:ablation}, show an evidently improved accuracy-latency tradeoff compared to the baseline, particularly at large pruning ratios and latency reduction(rightmost points in the curves). This indicates the effectiveness of "block grouping" even when used independently.

\vspace{-10pt}\subsubsection{}By integrating our above two strategies into a unified MINLP framework, we enable efficient and single-pass pruning. 

\begin{figure}[t!]
  \begin{minipage}[b]{0.49\textwidth}
    \centering
    \includegraphics[width=.85\linewidth]{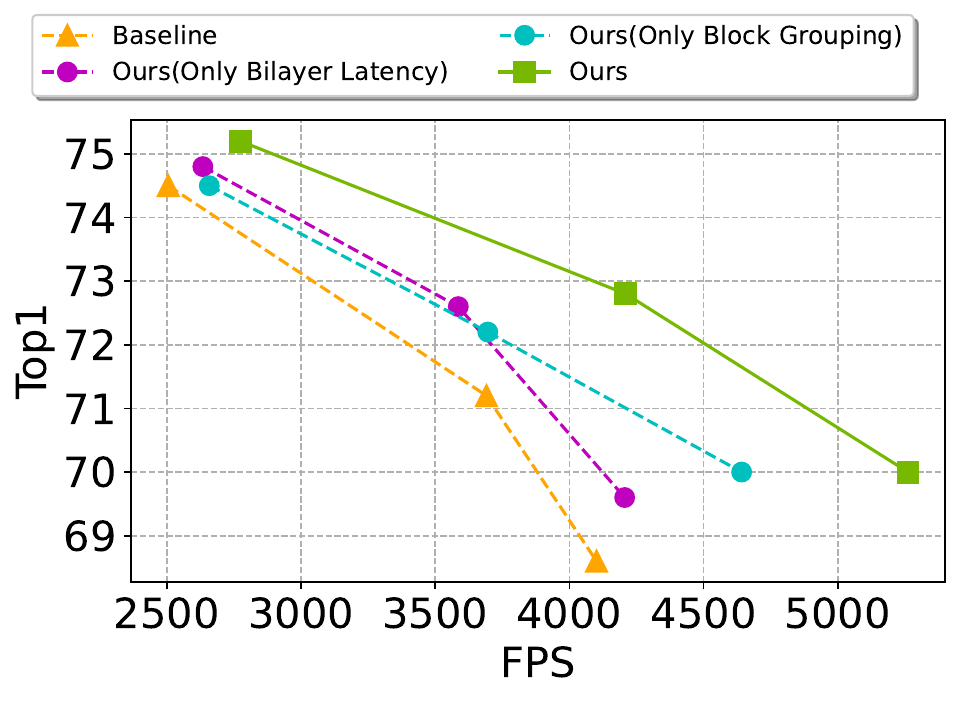}
    \vspace{-10pt}
    \captionof{figure}{\textbf{Ablation study results on ImageNet with ResNet50}. We show results of each improvement acting individually. Top-right is better.}
    \label{fig:ablation}
  \end{minipage}
  \hfill
  \begin{minipage}[b]{0.49\textwidth}
    \centering
    \resizebox{.9\linewidth}{!}
    {
    \begin{tabular}{cc|lc}
        \toprule
        \rowcolor{lgray}
        \textsc{Method} & \textsc{Steps} & \textsc{Top-1$\uparrow$}& \textsc{FPS$\uparrow$} \\
        \midrule
        \textsc{HALP-70\%} & $30$  & $74.3$ & $2505$ \\
        \textsc{HALP-70\%} & $10$  & $73.9$ & $2622$ \\
        \textsc{HALP-70\%} & $1$  & $65.1$ & $4444$ \\
        \hdashline
        \textsc{Ours$_{30}$-70\%} & $30$  & $74.5$ & $2660$ \\
        \textsc{Ours$_{10}$-70\%} & $10$  & $74.8$ & $2860$ \\
        \rowcolor{lgreen}
        \textbf{\textsc{Ours-70\%}} & $1$  & $75.2$ & $2774$ \\
        \bottomrule
        \end{tabular}
    }
      \captionof{table}{\textbf{Ablation study results on ImageNet with ResNet50}. We show results of ours and HALP~\cite{shen2021halp} with different pruning steps.}
      \label{table:ablation}
    \end{minipage}
    \vspace{-20pt}
\end{figure}

\noindent\textbf{Single-pass v.s. Iterative Pruning}
Our single-pass pruning approach achieves the target latency in just one step, while iterative methods like HALP~\cite{shen2021halp} require up to 30 steps. Performance comparisons between our method and HALP across different pruning steps are shown in Table~\ref{table:ablation}.

As observed, our approach performs consistently well regardless of the pruning steps. Our single-pass performance is even better than our 30-steps iterative pruning. We believe this is likely due to the benefit of using importance scores from all samples in the dataset at once.
 
In contrast, HALP's performance worsens with fewer pruning steps, especially in single-pass pruning where it defies the latency budget and over-prunes, leading to a Top-1 of $65.1$ and FPS of $4444$. This behavior is because the oversights of (a) and (b) can be somewhat mitigated with multiple pruning steps, but become more pronounced with just one or fewer steps.



\vspace{-8pt}
\section{Conclusion}
\vspace{-8pt}
In this paper, we introduce a novel pruning framework called \method{} that integrates channel, layer, and block pruning within a unified optimization process and develop an accurate latency modeling technique that captures simultaneous input and output channel variations. To incorporate these strategies, we reformulate pruning as a Mixed-Integer Nonlinear Program (MINLP) to efficiently identify the optimal pruned structure within a specific latency budget in a single pass. Our results demonstrate substantial improvements over previous methods, especially in scenarios requiring large pruning. We also provide an in-depth ablation study to investigate each contribution individually. 
%
%
\bibliographystyle{splncs04}
\bibliography{main}
\clearpage
\renewcommand*{\thesection}{\Alph{section}}
\setcounter{section}{0}
\textbf{\large Appendix}
\section{Detailed Analysis on Prior Latency Estimation}
In the main paper, we note that previous approaches~\cite{shen2021halp, shen2023hardware, humble2022soft} use an imprecise estimation of latency, resulting in suboptimal trade-offs between accuracy and latency. These methods consider changes in output channels but ignore concurrent variations in input channels that occur due to pruning the previous layer. This problem becomes more significant when targeting substantial reductions in latency. In this section, we will examine their limitations through detailed formulations.

To model the latency variations, previous methods compute a latency cost for the $j$-th channel at the $l$-th layer $\Theta_l^j$ as:
\begin{align}
\label{eqn: latency}
    R_l^j = T_l(p_{l-1}, j) - T_l(p_{l-1}, j-1)
\end{align}
Here, $T(.)$ is the pre-built lookup table, and $T_l(i, j)$ returns the latency of layer $l$ with $i$ input channels and $j$ output channels. Since previous methods leverage an iterative pruning technique to gradually reach the desired latency target, $p_{l-1}$ is the \textbf{total} number of input channels kept from the last pruning step. It is used as a constant input channel count in the current pruning step to query the latency from the lookup table $T_l$. However, we can not ascertain that the input channel count will remain the same throughout the current pruning step, which can only happen if previous layer is not pruned at all. The actual value of input channel count $\hat{p}_{l-1}$ will be smaller than this constant estimate $p_{l-1}$ if pruning takes place in the previous layer. Therefore, we have the following inequalities:
\begin{align}
    \hat{p}_{l-1} &\leq p_{l-1}\\
    T_l(\hat{p}_{l-1}, j) &\leq T_l(p_{l-1}, j)\\
    T_l(\hat{p}_{l-1}, j-1) &\leq T_l(p_{l-1}, j-1)
\end{align}

We can write the error of the latency estimation for each channel from the true value as:
\begin{align}
    \epsilon_l^j &= |\hat{R}_l^j - R_l^j| \\
    &= |T_l(\hat{p}_{l-1}, j) - T_l(\hat{p}_{l-1}, j-1) - T_l(p_{l-1}, j) + T_l(p_{l-1}, j-1)|\\
    &= |(T_l(p_{l-1}, j-1) - T_l(\hat{p}_{l-1}, j-1)) + (T_l(\hat{p}_{l-1}, j) - T_l(p_{l-1}, j))| \\
    &\leq |T_l(p_{l-1}, j-1) - T_l(\hat{p}_{l-1}, j-1)| + |T_l(p_{l-1}, j) + T_l(\hat{p}_{l-1}, j)|
\end{align}
When targeting large pruning ratio, the difference between $p_{l-1}$ and $\hat{p}_{l-1}$ will be enlarged since we expect to aggressively prune each layer and the input channel count will change substantially overtime. We could see from the formula that this latency modeling will be much more prone to make mistakes in measuring the channel latency cost under such scenarios.

In terms of our proposed latency estimation technique, instead of capturing latency cost for each channel, we directly model the latency for all configurations with differing input and output channel counts to enumerate the entire latency landscape at layer $l$. Recall our latency cost matrix $\mathbf{C}_l$ looks like:
\begin{align}
{
  \mathbf{C}_l =
  \left[ {\begin{array}{cccc}
    T_l(1,1) & T_l(1,2)& \cdots & T_l(1,m_l)\\
    T_l(2, 1) & T_l(2,2) & \cdots & T_l(2, m_l)\\
    \vdots & \vdots & \ddots & \vdots\\
    T_l(m_{l-1}, 1) & T_l(m_{l-1}, 2) & \cdots & T_l(m_{l-1}, m_l)\\
  \end{array} } \right]
}
\end{align}

Previous approaches could be viewed as \textbf{only leveraging the $p_{l-1}$th row of $\mathbf{C}_l$}:
\begin{align}
{
  \left[ {\begin{array}{cccc}
    T_l(p_{l-1},1) & T_l(p_{l-1},2)& \cdots & T_l(p_{l-1},m_l)\\
  \end{array} } \right]
}
\end{align}.

\section{Solving MINLPs}
In order to solve our MINLP program, we leverage the method called OA~\cite{duran1986outer,fletcher1994solving} which decomposes the problem into solving an alternating finite sequence of NLP subproblems and relaxed versions of MILP master program. We also leverage a method called Feasibility Pump~\cite{bonami2009feasibility} to   to expedite the process of finding feasible solutions within constraints. 

The entire program could be efficiently solved on common CPUs for modern network sizes. For instance, when applied to a model like ResNet50~\cite{he2016deep}, the entire optimization problem can be solved in approximately 5 seconds on an Intel Xeon E5-2698 CPU.

\begin{figure}[t!]
  \begin{minipage}[h]{.54\linewidth}
    \centering
    \includegraphics[width=\linewidth]{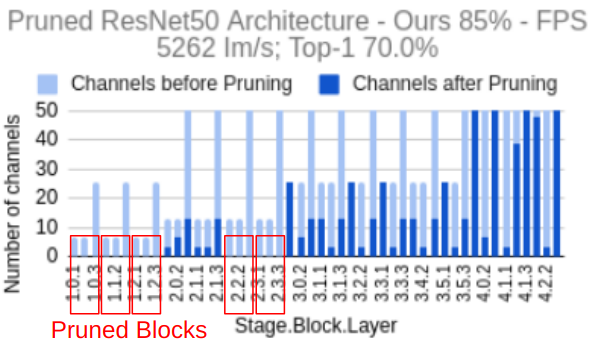}
    \vspace{-20pt}
    \captionof{figure}{Pruned architecture.}
    \label{fig:pruned_arch}
  \end{minipage}
  \hfill
  \begin{minipage}[h]{.44\linewidth}
    \centering
    \resizebox{\linewidth}{!}
    {
        \begin{tabular}{c|cc}
            \toprule
            \rowcolor{lgray}
            \textsc{Method} & \textsc{Top-1} & \textsc{FPS}\\
            \textsc{Autoslim}[\textcolor{eccvblue}{67}] & $74.0$  & $33.3$\\
            \textsc{EagleEye}[\textcolor{eccvblue}{36}] & $74.2$  & $31.3$\\
            \textsc{MetaPrune}[\textcolor{eccvblue}{43}] & $73.4$  & $33.3$\\
            \textsc{HALP}-$70\%$[\textcolor{eccvblue}{52}] & $74.5$  & $45.9$\\
            \rowcolor{lgreen}
            \textbf{Ours} & $\mathbf{75.2}$ & $\mathbf{118.2}$ \\
            \bottomrule
        \end{tabular}
    }
      \captionof{table}{Results on Intel CPU Xeon E5.}
      \label{table:cpu}
    \end{minipage}
    \vspace{-12pt}
\end{figure}

\section{Pruned Architecture} To gain deeper understanding of our pruning framework, we demonstrate the pruned structure in Fig.~\ref{fig:pruned_arch}. Our optimization approach maximizes global importance of a neural network under user-defined latency constraint (Eqn.\ref{eqn:program}), eliminating the need to pre-define pruning ratios. This automatically finds the optimal solution, pruning blocks only if their impact on overall importance is minimal compared to other options. The resulting pruned structure (Fig. \ref{fig:pruned_arch}) reveals that pruning is concentrated in shallower layers, contrary to the general expectation of layer collapse in deeper layers due to smaller gradients. 

\section{Performance on CPU Platform}
To demonstrate the adaptability and effectiveness of our approach across platforms, we instantiated and evaluated our method with the goal of optimizing latency on the Intel Xeon E5 CPU. The results, shown in Table \ref{table:cpu}, demonstrate significant performance improvements over previous work. Specifically, compared to HALP~\cite{shen2021halp}, we achieve higher Top-1 accuracy (75.2 vs. 74.5) while more than doubling the FPS (118.2 vs. 45.9). This improvement is even more pronounced than our GPU results presented in the main paper. We attribute this to our framework's ability to prune blocks, resulting in a shallower structure that is more conducive to CPU inference. This further underscores our method's potential for widespread applicability. Algorithmically, to enable different latency optimization targets (e.g., CPU vs. GPU), our framework only requires generating a latency lookup table $\mathcal{T}_l$ (\textit{Section 3.1 Latency Modeling}) on the target platform and feeding it into the MINLP program.

\section{Integration with Soft Masking}
Recent advances in pruning~\cite{humble2022soft, zhou2021learning, he2018soft, kusupati2020soft, kim2021dynamic} have increasingly adopted soft masking techniques to retain the learning capacity of pruned models by not directly removing the pruned weights. Notably, SMCP~\cite{humble2022soft} integrates this method into the HALP hardware-aware pruning framework, resulting in an enhanced accuracy-latency tradeoff for pruned models. Here, we explore the potential of soft masking to enhance our model's performance. 

\begin{figure}[h!]
\begin{subfigure}[b]{.55\textwidth}
\begin{center}
   \includegraphics[width=\linewidth]{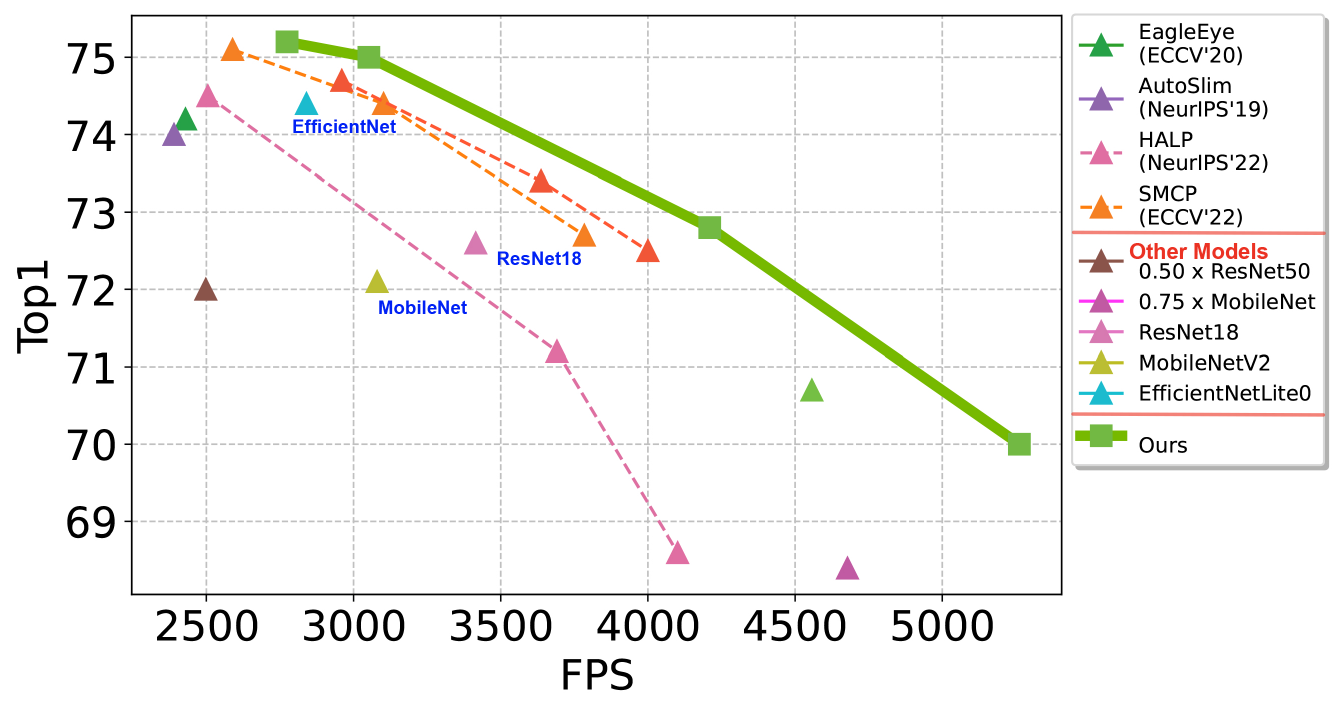}
\end{center}
\caption{\textbf{Comparison with smaller networks on ImageNet with pruning ResNet50}. Our approach of pruning large models across various ratios achieves a superior accuracy-speed trade-off compared to existing smaller networks. Top-right is better.}
\label{fig:smallarch}
\end{subfigure}
\hfill
\begin{subfigure}[b]{.45\textwidth}
\begin{center}
   \includegraphics[width=.9\linewidth]{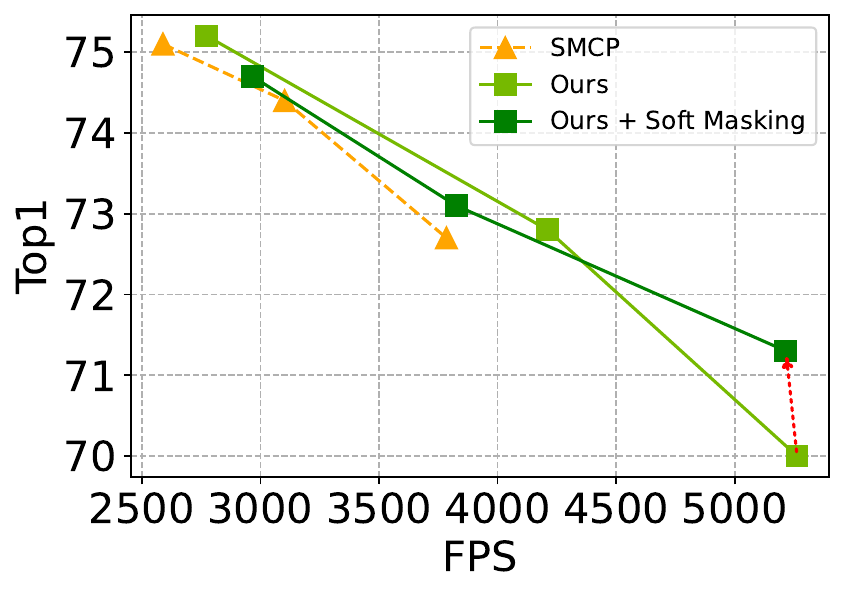}
\end{center}
\caption{\textbf{Results of ours with soft masking on ImageNet with ResNet50}. Improvement is observed in Top1 at a high FPS level. Top-right is better.}
\label{fig:softmaskg}
\end{subfigure}
\end{figure}

We conduct this study on ImageNet with ResNet50 and depict the Pareto frontier of FPS versus Top-1 in Figure~\ref{fig:softmaskg}. For clarity, we also include the performance of SMCP~\cite{humble2022soft} and ours. The results reveal that soft masking offers limited advantages at lower FPS levels with modest pruning ratios and latency reduction. Nonetheless, targeting higher FPS levels leads to notable improvements in Top-1 accuracy. This outcome may be attributed to the Taylor channel importance score we employed~\cite{molchanov2019importance}, which gauges parameter significance based on its impact on loss. Though it maintains precision with minor parameter deletions, its reliability may diminish when a larger number of parameters are pruned concurrently. The iterative reassessment inherent to the soft masking technique may counteract this shortcoming.

\section{Comparison with Smaller Networks} In our paper, we adopt a \textit{\textbf{"training large - pruning to small"}} paradigm. Our approach is particularly effective at large pruning ratios. Therefore, instead of starting with and pruning smaller architectures like MobileNet, EfficientNet, and ResNet18, we \textbf{\textit{aggressively}} prune a larger portion from ResNet50 to attain an efficient model. This strategy outperforms other smaller networks in both Top-1 accuracy and FPS, as illustrated in Figure~\ref{fig:smallarch}. Larger networks provide a richer manifold space for learning more complex features during training compared to smaller ones, giving the pruned models greater potential. Our framework excels at identifying an optimal pruned subnetwork structure, surpassing the performance of training smaller networks from scratch.

\section{Training Detail} The hyperparameters and settings used for experiments are detailed in Table \ref{tab:detail}.
\begin{table}[H]
    \centering
    \resizebox{.99\linewidth}{!}{
    \begin{tabular}{c|ccc}
        \toprule
        \rowcolor{lgray}
        Dataset & Epochs & Optimizer, Momentum, WeightDecay & Learning Rate\\
        ImageNet & $90$  & SGD, $0.875$, $3e-5$ &Init=$1.024$, LinearDecay\\
        PascalVOC & $800$ & SGD, $0.9$, $2e-3$ &Init=$8e-3$, StepDecay \\
        NuScenes & $60$  & AdamW,$n/a$, $0.01$ & Init=$6e-4$, CosineAnneal \\
        \bottomrule
    \end{tabular}
    }
    \caption{Training Detail.}
    \label{tab:detail}
\end{table}
\end{document}